\pdfoutput=1

\documentclass[11pt,usenames,dvipsnames]{article}
\usepackage{authblk}

\usepackage[]{acl}

\usepackage{times}
\usepackage{latexsym}
\usepackage{graphicx}
\usepackage{tikz}
\usetikzlibrary{matrix,positioning}

\usepackage[T2A, T1]{fontenc}

\usepackage[utf8]{inputenc}
\usepackage[english]{babel}
\usepackage{CJKutf8}

\usepackage{microtype}
\usepackage{booktabs}

\usepackage{tabularx}
\title{Construction Grammar Provides Unique Insight\\ into Neural Language Models}

\author[*$\diamond$]{Leonie Weissweiler}
\author[$\dag$]{Taiqi He}
\author[$\dag$]{Naoki Otani}
\author[$\dag$]{\\David R. Mortensen}
\author[$\dag$]{Lori Levin}
\author[*$\diamond$]{Hinrich Sch\"utze\vspace{-1em}}

\affil[*]{Center for Information and Language Processing, LMU Munich}
\affil[$\diamond$]{Munich Center of Machine Learning}
\affil[$\dag$]{Language Technologies Institute, Carnegie Mellon University
 \protect\\
\texttt{weissweiler@cis.lmu.de} \protect\\ \texttt{\{taiqih,notani,dmortens,lsl\}@cs.cmu.edu}}

\begin{document}
\maketitle
\begin{abstract}
Construction Grammar (CxG) has recently been used as the basis for probing studies that have investigated the performance of large pretrained language models (PLMs) with respect to the structure and meaning of constructions. In this position paper, we make suggestions for the continuation and augmentation of this line of research. We look at probing methodology that was not designed with CxG in mind, as well as probing methodology that was designed for specific constructions. We analyse selected previous work in detail, and provide our view of the most important challenges and research questions that this promising new field faces.

\end{abstract}

\section{Introduction}

In this paper, we will analyse existing literature investigating how well constructions and constructional information are represented in pretrained language models~(PLMs). We provide context to support the argument that this is one of the most important challenges facing Language Models~(LMs) today, and provide a summary of the current open research questions and how they might be tackled.

Our paper is organised as follows: In Section~\ref{motivation}, we explain why LMs must understand constructions to be good models of language and perform effectively on downstream tasks. In Section \ref{problem}, we analyse the existing literature on non-CxG-focused probing to determine its limitations in analysing constructional knowledge. In Section \ref{related}, we summarise the existing probing work that is specific to CxG and analyse its data, methodology, and findings. In Section \ref{research-questions}, we argue that the development of an appropriate probing methodology for constructions remains an open and important research question (\S\ref{methods}), and highlight the need for data collection and annotation for facilitating this area of research (\S\ref{data}). Finally, in Section \ref{training}, we suggest next steps that LMs might take if CxG probing reveals fundamental problems.

\subsection{Construction Grammar}
\label{cxg}

\begin{table*}[h]
\small
\centering
\begin{tabularx}{\linewidth}{lXX}

\toprule
\textbf{Construction Name}     & \textbf{Construction Template}                         
& \textbf{Examples} \\
\midrule

Word                   &                                & Banana                                                                        \\
Word (partially filled) &         pre-N, V-ing         &  Pretransition, Working                                                                                \\
Idiom (filled)      &                                   & Give the devil his due                                                     \\
Idiom (partially filled) &         Jog \textless{}someone's\textgreater{} memory       & She jogged his memory \\
Idiom (minimally filled) &The X-er the Y-er               & The more I think about it, the less I know                                         \\
Ditransitive construction (unfilled) & Subj V Obj1 Obj2  & He baked her a muffin                                                                        \\
Passive (unfilled)& Subj aux VPpp (PP by)               & The armadillo was hit by a car   \\
\bottomrule
    
\end{tabularx}
\caption{Standard examples of constructions at various levels, adapted from \citet{goldberghandbook}}
\end{table*}

\begin{figure}[t]
  \centering
  \begin{tikzpicture}
  \definecolor{myred}{RGB}{199, 47, 27}
\definecolor{myblue}{RGB}{49, 94, 216}
\definecolor{mygreen}{RGB}{90, 151, 33}
    \tikzset{
      fixed/.style = {rounded corners, thick, draw=myred, minimum height=0.6cm, text height=0.3cm, text depth=0.1cm},
      comparative/.style = {rounded corners, thick, draw=myblue, minimum height=0.6cm, text height=0.3cm, text depth=0.1cm},
      comparand/.style = {rounded corners, thick, draw=mygreen, minimum height=0.6cm, text height=0.3cm, text depth=0.1cm},
      lab/.style = {font=\tiny, text width=1.5cm, align=center}
    }
    \node[matrix] (mat) {
      \node[fixed] (the1) {the}; &
      \node[comparative] (funnier) {funnier}; &
      \node[comparand] (example) {the example};\\
      \node[minimum height=0.9cm] (dummy) {};
      \node[lab] (fixed) {fixed \textit{the}}; &
      \node[lab] (comparative) {comparative phrase}; &
      \node[lab] (comparands) {expressions being correlated}; \\
      \node[fixed] (the2) {the}; &
      \node[comparative] (more citations) {more citations}; &
      \node[comparand] (the paper) {the paper will have}; \\
  };
  \draw (example) -- (comparands) -- (the paper);
  \draw (the1) -- (fixed) -- (the2);
  \draw (funnier) -- (comparative) -- (more citations);

  \end{tikzpicture}
    \caption{An example illustrating the complexity of a construction. It is an instance of the English Comparative Correlative (CC), with its syntactic features highlighted above the text and paraphrases illustrating its meaning below.}
    \label{funny_example}
\end{figure}
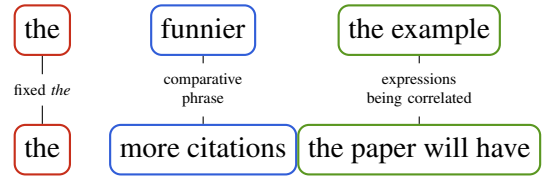

Although there are many varieties of CxG, they share the assumption that the basic building block of language structure is a pair of form and meaning. The form can be anything from a simple morpheme to the types of feature structures seen in Sign-Based Construction Grammar (SBCG)~\citep{Boas-Sag:2012}, which can be constellations of inflectional features, morphemes, categories like parts of speech, and syntactic mechanisms. Constructions with many detailed parts in SBCG include comparative constructions in sentences such as \textit{The desk is ten inches taller than the shelf}~\citep{HasegawaEtAl:2010} and the causal excess construction as in \textit{It was so big that it fell over}~\citep{Kay-Sag:2012}. Most importantly, the form or syntax of a sentence is not reduced to an idealized binary-branching tree or a set of hierarchically arranged pairs of head and dependants. For the purposes of this paper, we take the meaning of a construction to be a combination of Frame Semantics~\citep{ws-2014-frame} and comparative concepts in semantics and information packaging from language typology~\citep{Croft:2022}. Because CxG does not have a clear line separating the lexicon and the grammar, the same kinds of meanings that can be associated with words can be associated with more complex structures. Table~\ref{cxg}, adapted from \citet{goldberghandbook} illustrates constructions at different levels of complexity that contain different numbers of fixed lexemes and open slots. 

In this paper, we ask whether PLMs model constructions as gestalts in both form and meaning. For example, we want to know whether a PLM represents a construction like the Comparative Correlative (\textit{The more papers we write, the more fun we have}) as more than the sum of its individual phrases and dependencies. We also want to know whether the PLM encodes knowledge of the open slots in the construction and what can fill them. In terms of meaning, we want to find out whether the sentence's position in embedding space indicates that it has something to do with the correlation between the increase in writing more papers and having more fun. We would like to know whether PLMs represent the meaning of a correlative sentence as close to the meaning of other constructions in English and other languages that have different forms but similar meanings (e.g., \textit{When we write more papers, we have more fun}).


\subsection{Language Modelling}

This paper is partially concerned with the fundamental questions of language modelling: what is its objective, and what is required of a full language model? We see the objective of language modelling very pragmatically: we aim to build a system that can predict the words in a sentence as well as possible, and therefore our aim in this paper is to point out where this requires knowledge of constructions.
We do not take the objective of language modelling to mean that LMs should necessarily achieve their goal the same way that humans do. Therefore, we do not argue that language models need to ``think'' in terms of constructions because humans do. Rather, we consider constructions an inherent property of human language, which makes it necessary for language models to understand them. 


\section{Motivation}
\label{motivation}

There has recently been growing interest in developing probing approaches for PLMs based on CxG. We see these approaches as coming from two different motivational standpoints, summarised below.

\subsection{Constructions are Essential for Language Modelling}


According to CxG, meaning is encoded in abstract
constellations of linguistic units of different sizes. This means that LMs, which
the field of NLP is trying to develop to achieve human
language competency, must also be able to assign meaning to
these units to be full LMs. Their ability to assign
meaning to words, or more specifically to subword units
which are sometimes closer to morphemes than to words, has
been shown at length \cite{wiedemann2019,reif2019,schwartz-etal-2022-encode}. The question
therefore remains: are PLMs able to retrieve and use
meanings associated with patterns involving
multiple tokens? We do not take this to only mean contiguous, fixed expressions, but much more importantly, non-contiguous patterns with slots that have varying constraints placed on them. To imitate and match human language behaviour, models of human language need to learn how to recognise these patterns, retrieve their meaning, apply this meaning to the context, and use them when producing language. Simply put, there is no way around learning constructions if LMs are to advance. In addition, we believe that it is an independently interesting question whether existing PLMs pick up on these abstract patterns using the current architectures and training setups, and if not, which change in architecture would be necessary to facilitate this.

\subsection{Importance in Downstream Tasks}
\begin{table*}[h]
\centering
\small
\begin{tabularx}{\linewidth}{lXXX}
\toprule
\textbf{Lang} & \textbf{Reference Translation} & \textbf{DeepL Translation} \\ \midrule          
German & Sie nieste den Schaum von ihrem Cappuccino runter. & Sie nieste den Schaum von ihrem Cappuccino.   \\
Italian & Lei ha starnutito via la schiuma dal suo cappuccino. & Starnutì la schiuma del suo cappuccino. \\
Turkish & Cappuccino'sunun köpüğünü hapşırdı. & Hapşırarak cappuccino'sunun köpüğünü uçurdu. \\
\bottomrule
\end{tabularx}
\caption{Translations of `She sneezed the foam off her cappuccino.' given by DeepL\footnotemark[1]. 
Translated back to English by humans, they all mean ``She sneezed her cappuccino's foam.'', which does not correctly convey the resultative meaning component, i.e., that the foam is removed from the cappuccino by the sneeze (as opposed to put there).
}
\label{foam}
\end{table*}

Regardless of more fundamental questions about the long-term
goals of LMs, we also firmly believe that probing for CxG is
relevant for analysing the challenges that face applied NLP,
as evaluated on downstream tasks, at this point in
time. Discussion is increasingly focusing on diagnosing the
specific scenarios that are challenging for current
models. \citet{srivastava2022} propose test suites that are
designed to challenge LMs, and many of them are designed by
looking for `patterns' with a non-obvious, non-literal
meaning that is more than the sum of the involved words. One
example of such a failure can be found in Table \ref{foam},
where we provide the DeepL\footnote{\label{deepl}\url{https://www.deepl.com/translator}} translations for the famous
instance of the caused-motion construction
\citep[CMC;]{goldberg1995}: `She sneezed the foam off her
cappuccino', where the unusual factor is that \textit{sneeze} does
not usually take a patient argument or cause a motion. For
translation, this means that it either has to use the
corresponding CMC in the target language, which might be quite different in form from the English CMC, or paraphrase in a
way that conveys all meaning facets. For the languages we
tested,
DeepL did not achieve this: the resulting sentence sounds more like the foam was sneezed onto the cappuccino, or is ambiguous between this and the correct translation. Interestingly, for Russian, the motion is conveyed in the translation, but not the fact that it is caused by a sneeze.

Targeted adversarial test suites like this translation example can be a useful resource to evaluate how well LMs perform on constructions, but more crucially, CxG theory and probing methods will inform the design of better and more systematic test suites, which in turn will be used to improve LMs (\S\ref{training}).



\subsection{Diversity in Linguistics for NLP}

Discussions about PLMs as models of human language processing have recently gained popularity. One forum for such discussions is the Neural Nets for Cognition Discussion Group at CogSci2022\footnote{\url{http://neural-nets-for-cognition.net}}. The work is still very tentative, and most people agree that LMs are not ready to be used as models of human language processing. However, the discussion about whether LMs are ready to be used as cognitive models is dominated by results of probing studies based on Generative Grammar (GG), or more specifically Transformational Grammar. This means that GG is being used as the gold standard against which the cognitive plausibility of LMs is evaluated. Studies using GG assume a direct relationship between the models' performance on probing tasks and their linguistic competency. Increased performance on GG probing tasks is seen as a sign it is becoming more reasonable to use LMs as cognitive models. 
Another linguistic reason for theoretical diversity is that if we could show that LMs conform better to CxG rather than GG, this might open up interesting discussions if they ever start being used as cognitive models. 


\section{Established Probing Methods Are Only Applicable to Some Aspects of CxG} 
\label{problem}

Established probing methods have focused on different aspects of the syntactic and semantic knowledge of PLMs. In this section, we summarise the major approaches that were not designed specifically with constructions in mind. We show that although each of these methodologies deals with some aspect of CxG, and might even fully investigate some simpler constructions, none of them fully covers constructional knowledge as defined in Section~\ref{cxg}. 


\subsection{Probing Using Contextual Embeddings}


Various probing studies (\citealp{garcia-etal-2021-probing, chronis-erk-2020-bishop, karidi-etal-2021-putting, yaghoobzadeh-etal-2019-probing}; \textit{inter alia}) have focused on analysing contextual embeddings at different layers of PLMs, either of one word or multiple words, or both. The common thread in their methodology is that they compare the embeddings of the same word in different contexts, or of different words in the same context. From a constructional point of view, this requires finding two constructions with similar surface forms. By comparing the embeddings over many sentences, they are able to investigate if a certain word ``knows'' in which construction it is, which provides evidence for the constructional knowledge of a model. 

While this is a useful starting point for probing, it is also limited. Sentences with similar constructions have to be identified, which is not always possible. More importantly, this methodology currently does not tell us anything about if the model has identified the extent of the construction correctly, or if the model has correctly learned how each slot can be filled.

\subsection{Probing for Relationships Between Words}



Some probing studies investigate whether a PLM recognises a word pair associated with a meaningful relationship of some kind (\citet{rogers-etal-2020-primer}). Most prominently, probing based on Universal Dependencies (UD; \citet{de-marneffe-etal-2021-universal}) by \citet{hewitt-manning-2019-structural} attempts to find out whether there is a high attention weight between words that are in a dependency relation where one word is the head and the other word is the dependent. They found different attention heads at different layers that seem to represent specific dependency relations such as a direct object attending to its verb, a preposition attending to its object, determiners attending to nouns, possessive pronouns attending to head nouns, and passive auxiliary verbs attending to head verbs. 




The methodology as it was used by \citet{hewitt-manning-2019-structural} looked at the one token that each token attended to the most. This made sense for the \citet{hewitt-manning-2019-structural} study because they were probing for UD structures, which consist of binary relationships of heads and dependents in a hierarchical structure.

However, the methodology would have to be extended if we want to find out whether a whole construction with many construction elements is represented in the model in something other than a hierarchical set of binary relations. Most varieties of CxG recognise constructions with more than two daughters and constructions such as \textit{thirty miles an hour} \citep{fillmore-etal-constructicon} in which no element is the head (headless constructions). As a research question, it is still unclear what patterns of attention we would consider as evidence that a model encodes a construction that may have headless and non-binary branches. An appropriate probing methodology has not yet been developed. 




\subsection{Probing with Minimal Pairs}





Some works in probing based on Generative Grammar have relied on finding minimal pairs of sentences that are identical except for one
specific feature
that, if changed, will make the sentence ungrammatical
 \citep{wei-etal-2021-frequency}. For example, in \textit{The teacher who met the students is/*are smart}, a language model that encodes hierarchical structure would predict \textit{is} rather than \textit{are} after \textit{students}, whereas a language model that was fooled by adjacency might predict \textit{are} because it is next to \textit{students}. The sentences can be safely compared, because only one feature, in this case, the verb being assigned the same number as the subject, is changed, and no other information can intervene or distort the probe. Other studies use a more complicated paradigm of minimal pairs involving filler-gap constructions, contrasting \textit{I know what the lion attacked (gap) in the desert} and \textit{I know that the lion attacked the gazelle (no gap) in the desert}. 

These probing methodologies have led to productive lines of research and have been applied to complex constructions such as the Comparative Correlative Construction \cite{weissweiler-etal-2022-better}. However, they depend on finding two minimally different constructions, which differ only in one way (e.g., singular/plural or gap/no gap), but close minimal pairs are simply not available for every construction.

\section{CxG-specific Probing}
\label{related}

\begin{table*}[ht]
\centering
\small
\begin{tabularx}{\textwidth}{p{0.145\textwidth}lXXX}

\toprule
\textbf{Paper} & \textbf{Language} & \textbf{Source} & \textbf{Construction} & \textbf{Example}                                                  \\ \midrule
\citet{tayyar-madabushi-etal-2020-cxgbert} & English & From automatically constructed list by \citet{dunn2017} & Personal Pronoun + didn't + V + how & We didn't know how or why. \\[1cm]
\citet{li-etal-2022-neural}  & English & Argument Structure Constructions according to \citet{bencini2000}  &  caused-motion & Bob cut the bread into the pan. \\[1.3cm] 
\citet{tseng-etal-2022-cxlm}  & Chinese & From constructions list by \cite{zhan_2017} & \begin{CJK}{UTF8}{gbsn} a + 到
+ 爆\end{CJK}, etc. & \begin{CJK}{UTF8}{gbsn}好吃到爆了！\end{CJK}\newline \textit{It's so delicious!} \\[.7cm] 
\citet{weissweiler-etal-2022-better}  &  English & \citet{mccawley1988} & Comparative Correlative   & The bigger, the better. \\ 
\bottomrule
\end{tabularx}
\caption{Overview of constructions investigated in CxG-specific probing literature, with examples.}
\label{related_examples}
\end{table*}

We have argued that the most commonly used and straightforward probing methods are not sufficient for fully investigating constructional knowledge in PLMs.
However, there have been several papers which have created new probing 
methodologies specifically for constructions. In this section, we will analyse them in terms of

\begin{itemize}
\setlength\itemsep{0em}
    \item Which constructions were investigated? Does the paper investigate specific constructions or does it use a pre-compiled list of constructions or restrain itself to a subset?
    \item For the specific instances of their construction or constructions, what data are they using? Is it synthetic or collected from a corpus? If from a corpus, how was it collected?
    \item What are the key probing ideas?
    \item Does the paper only investigate probing of (unchanged) pretrained models or is
finetuning also considered?
\end{itemize}

For ease of reference, we provide an overview of the constructions investigated by each of the papers in Table \ref{related_examples}.



\subsection{CxGBERT}
\citet{tayyar-madabushi-etal-2020-cxgbert} investigate how well BERT \cite{devlin-etal-2019-bert} can classify whether two sentences contain instances of the same construction. Their list of constructions is extracted with a modified version of \citet{dunn2017}'s algorithm: they induce a CxG in an unsupervised fashion over a corpus, using statistical association measures. Their list of constructions is taken directly from \citet{dunn2017}, and they find their instances by searching for those constructions' occurrences in WikiText data. This makes the constructions possibly problematic, since they have not been verified by a linguist, which could make the conclusions drawn later from the results about BERT's handling of constructions hard to generalise from.

The key probing question of this paper is: Do two sentences
contain the same construction? This does not necessarily
need to be the most salient or overarching construction of
the sentence, so many sentences will contain more than one
instance of a construction. Crucially, the paper does not
follow a direct probing approach, but rather finetunes or
even trains BERT  on targeted construction data, to
then measure the impact on CoLA. They find that on average,
models trained on sentences that were sorted into documents
based on their constructions do not reliably perform better
than those trained on original, unsorted data. However, they
additionally test BERT Base with no additional pre-training
on the task of predicting whether two sentences contain instances
of the same construction, measuring accuracies of about 85\%
after 500 training examples for the probe. These results
vary wildly depending on the frequency of the construction,
which might relate back to the questionable quality of the
automatically identified list of constructions.

\subsection{Neural Reality of Argument Structure Constructions}
\citet{li-etal-2022-neural}
probe for LMs' handling of four argument structure constructions:
ditransitive, resultative, caused-motion, and
removal. Specifically, they attempt to adapt the findings
of \citet{bencini2000}, who used a sentence sorting task to
determine whether human participants perceive the argument
structure or the verb as the main factor in the overall sentence
meaning. The paper aims to recreate this experiment for
MiniBERTa \cite{warstadt-etal-2020-learning} and
RoBERTa \cite{liu2019roberta}, by generating sentences
artificially and using agglomerative clustering on the
sentence embeddings. They find that, similarly to the human
data, which is sorted by the English proficiency of the
participants, PLMs increasingly prefer sorting by
construction as their training data size
increases. Crucially, the sentences constructed for testing
had no lexical overlap, such that this sorting preference
must be due to an underlying recognition of a shared pattern
between sentences with the same argument structure. They
then conduct a second experiment, in which they insert
random verbs, which are incompatible with one of the
constructions, and then measure the Euclidean distance
between this verb's contextual embedding and that of a verb
that is prototypical for the corresponding construction. The
probing idea here is that if construction information is
picked up by the model, the contextual embedding of the verb
should acquire some constructional meaning, which would
bring it closer to the corresponding prototypical verb
meaning than to the others. They indeed find that this
effect is significant, for both high and low frequency
verbs.


\subsection{CxLM}

\citet{tseng-etal-2022-cxlm}
study LM predictions for the slots of various degrees of
openness for a corpus of Chinese constructions. Their
original data comes from a knowledge database of Mandarin
Chinese constructions \cite{zhan_2017}, which they filter so
that only constructions with a fixed repetitive element
remain, which are easier to find automatically in a
corpus. They filter this list down further to constructions
which are rated as commonly occurring by annotators, and
retrieve instances from a POS-tagged 
Taiwanese bulletin board corpus.
They binarise the openness of a given slot in a
construction and mark each word in a construction as either
constant or variable. The key probing idea is then to
examine the conditional probabilities that a model outputs
for each type of slot, with the expectation that the
prediction of variable slot words will be more difficult
than that of constant ones, providing that the model has
acquired some constructional knowledge. They find that this
effect is significant for two different Chinese BERT-based
models, as negative log-likelihoods are indeed significantly
higher when predicting variable slots compared to constant
ones. Interestingly, the
negative log-likelihood resulting from masking the
entire construction lies in the middle of the two
extremes. They further evaluate  a BERT-based model which
is finetuned on just predicting the variable slots of the
dataset they compiled and find, unsurprisingly, that this
improves accuracy greatly.

\subsection{Probing for the English Comparative Correlative}

\citet{weissweiler-etal-2022-better}
investigate large PLM performance on the English Comparative
Correlative (CC). There are two key probing ideas,
corresponding to the investigation of the syntactic vs.\ the
semantic component of CC.
They probe
for PLM understanding of CC's syntax by attempting
to create minimal pairs, which consist of sentences with
instances of the CC and very similar sentences which do not
contain an instance of the CC. They collect minimal pairs
from data by searching for sentences that fit the general
pattern and manually annotate them as positive and
negative instances, and additionally construct artificial
minimal pairs that turn a CC sentence into a non-CC sentence by reordering words. They find that a probing
classifier can distinguish between the two classes easily,
using mean-pooled contextual PLM embeddings.
They also probe
the models' understanding of the meaning of CC, for which they choose a usage-based approach,
constructing NLU-style test sentences in which an instance
of the construction is given and has then to be applied in a
context. They find no above-chance performance for any of
the models investigated in this task.

\subsection{Summary}

In this section, we summarise the findings of previous work
on CxG-based LM probing and analyse them in terms of the
constructions that are investigated, the data that is used
and the probing approaches that are applied.

\subsubsection{Constructions Used}

So far, \citeposs{tseng-etal-2022-cxlm} study is only the work that chose a set of constructions from a list precompiled by linguists. They constrain their selection to
contain only constructions that are easy to search for in a
corpus, and the resource they use only contains
constructions with irregular syntax, but it is
nevertheless to be considered a positive point that they are
able to reach a diversity of constructions investigated. In
contrast, both \citet{li-etal-2022-neural} and \citet{weissweiler-etal-2022-better}
pick one or a few constructions manually, both of which are
instances of `typical' constructions frequently discussed in
the linguistic literature.
This makes the work  more interesting to linguists and
the validity of the constructions is beyond doubt.
But the downside is selection bias: the constructions that
are frequently discussed are likely to have strong
associated meanings and do not constitute a representative
sample of constructions, from a
constructions-all-the-way-down
standpoint \cite{Goldberg.2006}. Lastly, \citet{tayyar-madabushi-etal-2020-cxgbert}
rely on artificial data collected by \citet{dunn2017}.
We consider this method to be unreliable, but it has the
resulting dataset has the advantage
of variety and large scale.

\subsubsection{Data Used}

The two main approaches to collecting data are:
(i) \emph{patterns}: finding instances of
the constructions using patterns of words / part-of-speech (POS) tags
and (ii) \emph{generation}  of synthetic data.
\citet{tseng-etal-2022-cxlm}, \citet{weissweiler-etal-2022-better}
and \citet{tayyar-madabushi-etal-2020-cxgbert}
use patterns
while \citet{li-etal-2022-neural} and a part
of \citet{weissweiler-etal-2022-better}
generate data based on
formal grammars. Patterns have
the advantage of natural data and are less prone to
accidental unwanted correlations. But there is a risk
of  errors in the data collection process, even
after the set of constructions has to be constrained to even
allow for automatic classification, and the data may have
been post-corrected by manual annotation, which is
time-intensive. On the other hand, generation
 bears challenges for making the sentences as natural as
possible, which can
eliminate
confounding factors like lexical overlap.

\subsubsection{Probing Approaches Used}

Regarding the probing approaches, all previous work has had
its own idea. \citet{weissweiler-etal-2022-better}
and \citet{li-etal-2022-neural} both operate on the level of
sentence embeddings, classifying and clustering them
respectively. \citet{tayyar-madabushi-etal-2020-cxgbert}
could maybe be classified with them, as it employs the Next Sentence Prediction
objective \cite{devlin-etal-2019-bert}, which operates at the sentence level. On the
other hand, another part of \citet{weissweiler-etal-2022-better}, as well
as \citet{tseng-etal-2022-cxlm}, works at the level of
individual predictions for masked tokens.

The greatest difference between these works is in their concept of evidence for constructional information learned by a model, and what this information even consists of.
\citet{tayyar-madabushi-etal-2020-cxgbert} frame this information as `do these two sentences contain the same construction', \citet{li-etal-2022-neural} as `is clustering by the construction preferred over clustering by the verb', \citet{weissweiler-etal-2022-better} as `can a small classifier distinguish this construction from similar-looking sentences' and `can information given in form of a construction be applied in context', and \citet{tseng-etal-2022-cxlm} as `are open slots more difficult to predict than closed ones'. There is little overlap to be found between these approaches, so it is difficult to draw any conclusion from more than one paper at a time.

\subsubsection{Overall Findings}
\label{general}

We nonetheless make an attempt at summarising the findings so far about large PLMs' handling of constructional information. Regarding the structure, all findings seem to be consistent with the idea that models have picked up on the syntactic structure of constructions and recognised similarities between different instances of the same construction. This appears to hold true even when tested in different rigorous setups that exclude bias from overlapping vocabulary or accidentally similar sentence structure. This has mostly been found for English, as \citet{tseng-etal-2022-cxlm} are the only ones investigating it for a non-English language, and it remains to be seen if it holds true for lower-resources languages. Considering the acquisition of the meaning of constructions, only \citet{weissweiler-etal-2022-better} have investigated this, and found no evidence that models have formed any understanding of it, but were not able to provide conclusive evidence to the contrary.

\section{Research Questions}\label{research-questions} 

In this section, we lay out our view of the problems that are facing the emerging field of CxG-based probing and the reasons behind these challenges, and propose avenues for potential future work and improvement.

\subsection{How Can We Develop Probing Methods that are a Better Fit for CxG?} 
\label{methods}

Going forward, we see two directions. One is what has
already been happening: keep finding new ways to
get around the inherent difficulty of probing for
constructions, which leads us to mostly non-conclusive and
not entirely reliable evidence. The better, and more
difficult way forward, is to adopt a fundamentally
different methodology that would establish a
standard of evidence/generalisability comparable to
GG-based probing.

\subsection{Data}
\label{data}

Another reason why so little work has been done in this important field is likely the lack of data. We view the lack of data as divided into three parts: the lack of lists of constructions, the lack of meaning descriptions or even a unified meaning formalism for them, and the lack of annotated instances in corpora. We explain different opportunities for the community to obtain this data going forward below. 

\subsubsection{Exploiting Non-constructicon Data}
Many resources are available, as already stated above, that have collected or created data with specific constructions, with the aim of making certain tasks more challenging to the models in a specific way. We can analyse those datasets and the results on them from a CxG point of view, and this can add to our pool of knowledge about what models struggle with regarding constructions. They will probably not contain any meaning descriptions, but some, like in \citet{srivastava2022}, are grouped naturally by construction, and contain instances in data, which may however be artificial. 

\subsubsection{Making Constructicons Available}

Recently, there has been substantial work by linguists to
develop constructicons for different
languages \citep{lyngfelt2018,constructicon}. Some of these
constructicons are readily available online,
e.g., the Brazilian Portuguese one, but many are either not
available or have an interface that makes them difficult to
access, e.g.,  because it is in the constructicon's
language. Although to our knowledge, none of these
constructicons contain annotated instances in text, and
their meaning representations will be very difficult to
unify, they are an important resource at least for lists of
constructions that can be investigated by probing
methods. They are especially valuable because of their
linguistic diversity (English, German, Japanese, Swedish, Russian, Brazilian Portuguese), the lack of which is a major flaw in the current literature, as we stated above in \S\ref{general}. 

\subsubsection{Universal Constructicon}

As a more ambitious project than simply making these constructicons available online, we firmly believe that the field would benefit greatly from an attempt to unify their representations and make them available as a shared resource. Parallels can be drawn here to UD \citep{de-marneffe-etal-2021-universal}, a project which developed a simplified version of dependency syntax that could be universally applied and agreed upon, and then provided funding for the creation of initial resources for a range of languages, which was later greatly added to by community work in the different communities. This was a major factor in the popularisation of dependency syntax within the NLP community, to the point where it is now almost synonymous with syntax itself, due in no small part to its convenience for computational research. 

As a second step after the creation of a shared online resource to access the existing constructicons, the community could consider developing a shared representation to formalise the surface form of the constructions. A dataset without meaning representation that includes multiple languages would already be a very useful resource. As a next step after that, we could think about aligning constructions across languages that encode a similar meaning. The last and most ambitious step would be unifying and linking the meaning representations, which would ideally be formalised similarly to AMR  \cite{banarescu-etal-2013-abstract}. This would enable us to develop automatic test suites that can really account for the constructions' meanings and not just their structure.

\subsubsection{Annotated Instances in Text}

In any stage of the development of 'construction lists'
detailed above, it would be necessary to find instances of
the constructions in text. Some of the probing literature
described above have generated this data artificially, which
is time-consuming and also removes two important advantages
of precompiled construction lists: objectivity and
scale. Therefore, the ideal solution would be to find
resources to have data annotated for constructions. This in
itself faces many challenges from a
constructions-all-the-way-down perspective: annotating even
one sentence completely would be very time-consuming and
require many discussions about annotation schemata in
advance. A more basic way of acquiring data would be to
focus on a limited set of constructions, which is selected
manually, and to use pre-filtering methods similar to those
employed by \citet{tseng-etal-2022-cxlm}
and \citet{weissweiler-etal-2022-better}, to acquire simply an Inside-Outside-Beginning
marking in sentences that might be instances of a
construction. On the downside, this is far less
linguistically rigorous and also less timeless than
Universal Dependencies, which guarantees that any annotated
sentence has been fully annotated and will probably not need
to be revised. Nevertheless, a compromise will need to be
found if annotated data is to be created at all.

\subsection{CxG and Transformer Architecture}

As more work is done on CxG-based probing, the field will
hopefully soon be able to approach the questions that we see
as crucial.  Current probing techniques have not yet shown that PLMs are able to adequately handle the meaning of constructions. Assuming that more comprehensive probing techniques will show conclusively that this is not the case, is it due to a lack of data?
Or is there a fundamental incompatibility of current
architectures and the concept of associating a pattern with
a meaning?
In \ref{speculation1} and \ref{speculation2}, we elaborate on why the latter might be the case.

\subsubsection{Non-compositional Meaning}
\label{speculation1}

It is possible that constructions are intrinsically difficult for LMs because they include non-compositional meaning that is not attached to a token. 
It is tempting to compare them to simpler multiword expressions, which also have meaning that spans several words and that is only instantiated when they appear together. 
They also pose a challenge to LMs because of this, as their concept of sentence meaning is often too compositional \cite{liu2022}.
The key difference is in our view, that for very complex constructions, it is not clear where in the model we can search or probe for the additional meaning.

The meaning is not attached to the words instantiating the construction, but rather to the abstract pattern itself \cite{croft2001}, which we can recognise, connect mentally to previous instances and store meaning for. 
Once we have retrieved this meaning, it is potentially applied to the whole sentence, and can therefore have consequences for the contextual meaning of words which were never even involved in it. 
In a transformer-based LM, this additional meaning component cannot be stored in the static embeddings and contextualised through the attention layers, because unlike for MWEs, many constructions have very open slots, so that it is impossible to say that their meaning should somehow be stored with the meaning of the words that may instantiate them. 
The only place to store constructional information, therefore, remains the model weights, which are much harder to investigate or alter than the model's input, and further probing might reveal that they are unable to store it at all.

\subsubsection{The Language Modelling Objective}
\label{speculation2}

Another possibility for fundamental difficulties arises from the nature of the training objective. PLMs are typically trained either on a masked or causal language modelling objective~\cite{devlin-etal-2019-bert,gpt2}. It makes sense that this incentivises them to learn word meaning in context, which they will need to predict certain words, and also relationships between words, such as simple morphological dependencies. However, information about the meaning of a construction might not often be learned in a language modelling setting, simply because it will not be needed to make the correct prediction. The meaning of a construction might not be necessary information to predict one of its component words correctly when it is masked, although its structure certainly will. 
In contrast, finetuning on a downstream task that requires assessment of sentence meaning, such as sentence classification, might enable us to better access the constructional meaning contained in PLMs, because the finetuning objective has required explicit use of this meaning. On the other hand, this might also be thought of as a distortion of the lens, as grammatical knowledge is not typically evaluated on finetuned models, because the findings might not generalise well.

\subsection{Adapting Pretraining for CxG}
\label{training}
If we do decide that there is a fundamental problem with the current architecture and/or training regime, the next logical step would be to think about how to alter these so that acquisition of constructional meaning becomes possible. Something similar has already been considered by \citet{tseng-etal-2022-cxlm}, where models are finetuned on data that has been altered to mask entire construction instances at once, and by \citet{tayyar-madabushi-etal-2020-cxgbert}, which collects sentences that contain instances of the same construction into `documents' and pretrains on them. 
This line of thinking, which can be summarised as data modification with constructional biases, can be further expanded, to give models some help with associating sentences with similar constructions with each other.

A far more radical idea would be to think about injecting something into the architecture that could represent this additional meaning, in the style of a position embedding, or a control token \cite{martin-etal-2020-controllable}.



\section{Conclusion}

We have motivated why probing large PLMs for CxG is a very important topic both for computational linguists interested in the ideal LM and for applied NLP scientists seeking to analyse and improve the current challenges that models are facing. We then summarised and analysed the existing literature on this topic. Finally, we have given our reasons for why CxG probing remains a challenge, and detailed suggestions for further development in this field, within the realms of data, methodology, and fundamental research questions.

\bibliography{anthology,custom}

\begin{thebibliography}{39}
\expandafter\ifx\csname natexlab\endcsname\relax\def\natexlab#1{#1}\fi

\bibitem[{Banarescu et~al.(2013)Banarescu, Bonial, Cai, Georgescu, Griffitt,
  Hermjakob, Knight, Koehn, Palmer, and
  Schneider}]{banarescu-etal-2013-abstract}
Laura Banarescu, Claire Bonial, Shu Cai, Madalina Georgescu, Kira Griffitt, Ulf
  Hermjakob, Kevin Knight, Philipp Koehn, Martha Palmer, and Nathan Schneider.
  2013.
\newblock \href {https://aclanthology.org/W13-2322} {{A}bstract {M}eaning
  {R}epresentation for sembanking}.
\newblock In \emph{Proceedings of the 7th Linguistic Annotation Workshop and
  Interoperability with Discourse}, pages 178--186, Sofia, Bulgaria.
  Association for Computational Linguistics.

\bibitem[{Bencini and Goldberg(2000)}]{bencini2000}
Giulia~ML Bencini and Adele~E Goldberg. 2000.
\newblock The contribution of argument structure constructions to sentence
  meaning.
\newblock \emph{Journal of Memory and Language}, 43(4):640--651.

\bibitem[{Boas and Sag(2012)}]{Boas-Sag:2012}
H.~C. Boas and I.~A. Sag. 2012.
\newblock \emph{Sign-Based Construction Grammar}.
\newblock Center for the Study of Language and Information.

\bibitem[{Chronis and Erk(2020)}]{chronis-erk-2020-bishop}
Gabriella Chronis and Katrin Erk. 2020.
\newblock \href {https://doi.org/10.18653/v1/2020.conll-1.17} {When is a bishop
  not like a rook? when it{'}s like a rabbi! multi-prototype {BERT} embeddings
  for estimating semantic relationships}.
\newblock In \emph{Proceedings of the 24th Conference on Computational Natural
  Language Learning}, pages 227--244, Online. Association for Computational
  Linguistics.

\bibitem[{Croft(2001)}]{croft2001}
William Croft. 2001.
\newblock \emph{Radical construction grammar: Syntactic theory in typological
  perspective}.
\newblock Oxford University Press on Demand.

\bibitem[{Croft(2022)}]{Croft:2022}
William Croft. 2022.
\newblock \emph{Morphosyntax: Constructions of the World's Languages}.
\newblock Cambridge University Press.

\bibitem[{de~Marneffe et~al.(2021)de~Marneffe, Manning, Nivre, and
  Zeman}]{de-marneffe-etal-2021-universal}
Marie-Catherine de~Marneffe, Christopher~D. Manning, Joakim Nivre, and Daniel
  Zeman. 2021.
\newblock \href {https://doi.org/10.1162/coli_a_00402} {{U}niversal
  {D}ependencies}.
\newblock \emph{Computational Linguistics}, 47(2):255--308.

\bibitem[{Devlin et~al.(2019)Devlin, Chang, Lee, and
  Toutanova}]{devlin-etal-2019-bert}
Jacob Devlin, Ming-Wei Chang, Kenton Lee, and Kristina Toutanova. 2019.
\newblock \href {https://doi.org/10.18653/v1/N19-1423} {{BERT}: Pre-training of
  deep bidirectional transformers for language understanding}.
\newblock In \emph{Proceedings of the 2019 Conference of the North {A}merican
  Chapter of the Association for Computational Linguistics: Human Language
  Technologies, Volume 1 (Long and Short Papers)}, pages 4171--4186,
  Minneapolis, Minnesota. Association for Computational Linguistics.

\bibitem[{Dunn(2017)}]{dunn2017}
Jonathan Dunn. 2017.
\newblock Computational learning of construction grammars.
\newblock \emph{Language and cognition}, 9(2):254--292.

\bibitem[{Fillmore et~al.(2012)Fillmore, Lee-Goldman, and
  Rhodes}]{fillmore-etal-constructicon}
C.~J. Fillmore, R.~Lee-Goldman, and R.~Rhodes. 2012.
\newblock The framenet constructicon.
\newblock In H.~C. Boas and I.~A. Sag, editors, \emph{Sign-Based Construction
  Grammar}. Center for the Study of Language and Information.

\bibitem[{Garcia et~al.(2021)Garcia, Kramer~Vieira, Scarton, Idiart, and
  Villavicencio}]{garcia-etal-2021-probing}
Marcos Garcia, Tiago Kramer~Vieira, Carolina Scarton, Marco Idiart, and Aline
  Villavicencio. 2021.
\newblock \href {https://doi.org/10.18653/v1/2021.eacl-main.310} {Probing for
  idiomaticity in vector space models}.
\newblock In \emph{Proceedings of the 16th Conference of the European Chapter
  of the Association for Computational Linguistics: Main Volume}, pages
  3551--3564, Online. Association for Computational Linguistics.

\bibitem[{Goldberg(2006)}]{Goldberg.2006}
Adele Goldberg. 2006.
\newblock \emph{Constructions at work: The nature of generalization in
  language}.
\newblock {Oxford University Press}, Oxford, UK.

\bibitem[{Goldberg(1995)}]{goldberg1995}
Adele~E.. Goldberg. 1995.
\newblock \emph{Constructions: A construction grammar approach to argument
  structure}.
\newblock University of Chicago Press.

\bibitem[{Goldberg(2013)}]{goldberghandbook}
Adele~E. Goldberg. 2013.
\newblock \href {https://doi.org/10.1093/oxfordhb/9780195396683.013.0002}
  {{1415 Constructionist Approaches}}.
\newblock In \emph{{The Oxford Handbook of Construction Grammar}}. Oxford
  University Press.

\bibitem[{Hasegawa et~al.(2010)Hasegawa, Lee-Goldman, Ohara, Fujii, and
  Fillmore}]{HasegawaEtAl:2010}
Yoko Hasegawa, Russell Lee-Goldman, Kyoko~Hirose Ohara, Seiko Fujii, and
  Charles~J Fillmore. 2010.
\newblock On expressing measurement and comparison in english and japanese.
\newblock \emph{Contrastive studies in construction grammar}, 10.

\bibitem[{Hewitt and Manning(2019)}]{hewitt-manning-2019-structural}
John Hewitt and Christopher~D. Manning. 2019.
\newblock \href {https://doi.org/10.18653/v1/N19-1419} {{A} structural probe
  for finding syntax in word representations}.
\newblock In \emph{Proceedings of the 2019 Conference of the North {A}merican
  Chapter of the Association for Computational Linguistics: Human Language
  Technologies, Volume 1 (Long and Short Papers)}, pages 4129--4138,
  Minneapolis, Minnesota. Association for Computational Linguistics.

\bibitem[{Karidi et~al.(2021)Karidi, Zhou, Schneider, Abend, and
  Srikumar}]{karidi-etal-2021-putting}
Taelin Karidi, Yichu Zhou, Nathan Schneider, Omri Abend, and Vivek Srikumar.
  2021.
\newblock \href {https://doi.org/10.18653/v1/2021.emnlp-main.806} {Putting
  words in {BERT}{'}s mouth: Navigating contextualized vector spaces with
  pseudowords}.
\newblock In \emph{Proceedings of the 2021 Conference on Empirical Methods in
  Natural Language Processing}, pages 10300--10313, Online and Punta Cana,
  Dominican Republic. Association for Computational Linguistics.

\bibitem[{Kay and Sag(2012)}]{Kay-Sag:2012}
Paul Kay and Ivan~A. Sag. 2012.
\newblock Cleaning up the big mess: Discontinuous dependencies and complex
  determiners.
\newblock In H.~C. Boas and I.~A. Sag, editors, \emph{Sign-Based Construction
  Grammar}. Center for the Study of Language and Information.

\bibitem[{Li et~al.(2022)Li, Zhu, Thomas, Rudzicz, and
  Xu}]{li-etal-2022-neural}
Bai Li, Zining Zhu, Guillaume Thomas, Frank Rudzicz, and Yang Xu. 2022.
\newblock \href {https://doi.org/10.18653/v1/2022.acl-long.512} {Neural reality
  of argument structure constructions}.
\newblock In \emph{Proceedings of the 60th Annual Meeting of the Association
  for Computational Linguistics (Volume 1: Long Papers)}, pages 7410--7423,
  Dublin, Ireland. Association for Computational Linguistics.

\bibitem[{Liu and Neubig(2022)}]{liu2022}
Emmy Liu and Graham Neubig. 2022.
\newblock Are representations built from the ground up? an empirical
  examination of local composition in language models.
\newblock \emph{arXiv preprint arXiv:2210.03575}.

\bibitem[{Liu et~al.(2019)Liu, Ott, Goyal, Du, Joshi, Chen, Levy, Lewis,
  Zettlemoyer, and Stoyanov}]{liu2019roberta}
Yinhan Liu, Myle Ott, Naman Goyal, Jingfei Du, Mandar Joshi, Danqi Chen, Omer
  Levy, Mike Lewis, Luke Zettlemoyer, and Veselin Stoyanov. 2019.
\newblock Roberta: A robustly optimized bert pretraining approach.
\newblock \emph{arXiv preprint arXiv:1907.11692}.

\bibitem[{Lyngfelt et~al.(2018)Lyngfelt, Borin, Ohara, and
  Torrent}]{lyngfelt2018}
Benjamin Lyngfelt, Lars Borin, Kyoko Ohara, and Tiago~Timponi Torrent. 2018.
\newblock \emph{Constructicography: Constructicon development across
  languages}, volume~22.
\newblock John Benjamins Publishing Company.

\bibitem[{Martin et~al.(2020)Martin, de~la Clergerie, Sagot, and
  Bordes}]{martin-etal-2020-controllable}
Louis Martin, {\'E}ric de~la Clergerie, Beno{\^\i}t Sagot, and Antoine Bordes.
  2020.
\newblock \href {https://aclanthology.org/2020.lrec-1.577} {Controllable
  sentence simplification}.
\newblock In \emph{Proceedings of the Twelfth Language Resources and Evaluation
  Conference}, pages 4689--4698, Marseille, France. European Language Resources
  Association.

\bibitem[{McCawley(1988)}]{mccawley1988}
James~D McCawley. 1988.
\newblock The comparative conditional construction in english, german, and
  chinese.
\newblock In \emph{Annual Meeting of the Berkeley Linguistics Society},
  volume~14, pages 176--187.

\bibitem[{Petruck and de~Melo(2014)}]{ws-2014-frame}
Miriam R.~L. Petruck and Gerard de~Melo, editors. 2014.
\newblock \href {https://doi.org/10.3115/v1/W14-30} {\emph{Proceedings of Frame
  Semantics in {NLP}: A Workshop in Honor of Chuck {F}illmore (1929-2014)}}.
  Association for Computational Linguistics, Baltimore, MD, USA.

\bibitem[{Radford et~al.(2019)Radford, Wu, Child, Luan, Amodei, Sutskever
  et~al.}]{gpt2}
Alec Radford, Jeffrey Wu, Rewon Child, David Luan, Dario Amodei, Ilya
  Sutskever, et~al. 2019.
\newblock Language models are unsupervised multitask learners.
\newblock \emph{OpenAI blog}, 1(8):9.

\bibitem[{Reif et~al.(2019)Reif, Yuan, Wattenberg, Viegas, Coenen, Pearce, and
  Kim}]{reif2019}
Emily Reif, Ann Yuan, Martin Wattenberg, Fernanda~B Viegas, Andy Coenen, Adam
  Pearce, and Been Kim. 2019.
\newblock \href
  {https://proceedings.neurips.cc/paper/2019/file/159c1ffe5b61b41b3c4d8f4c2150f6c4-Paper.pdf}
  {Visualizing and measuring the geometry of bert}.
\newblock In \emph{Advances in Neural Information Processing Systems},
  volume~32. Curran Associates, Inc.

\bibitem[{Rogers et~al.(2020)Rogers, Kovaleva, and
  Rumshisky}]{rogers-etal-2020-primer}
Anna Rogers, Olga Kovaleva, and Anna Rumshisky. 2020.
\newblock \href {https://doi.org/10.1162/tacl_a_00349} {A primer in
  {BERT}ology: What we know about how {BERT} works}.
\newblock \emph{Transactions of the Association for Computational Linguistics},
  8:842--866.

\bibitem[{Schwartz et~al.(2022)Schwartz, Haley, and
  Tyers}]{schwartz-etal-2022-encode}
Lane Schwartz, Coleman Haley, and Francis Tyers. 2022.
\newblock \href {https://aclanthology.org/2022.fieldmatters-1.8} {How to encode
  arbitrarily complex morphology in word embeddings, no corpus needed}.
\newblock In \emph{Proceedings of the first workshop on NLP applications to
  field linguistics}, pages 64--76, Gyeongju, Republic of Korea. International
  Conference on Computational Linguistics.

\bibitem[{Srivastava et~al.(2022)Srivastava, Rastogi, Rao, Shoeb, Abid, Fisch,
  Brown, Santoro, Gupta, Garriga-Alonso et~al.}]{srivastava2022}
Aarohi Srivastava, Abhinav Rastogi, Abhishek Rao, Abu Awal~Md Shoeb, Abubakar
  Abid, Adam Fisch, Adam~R Brown, Adam Santoro, Aditya Gupta, Adri{\`a}
  Garriga-Alonso, et~al. 2022.
\newblock Beyond the imitation game: Quantifying and extrapolating the
  capabilities of language models.
\newblock \emph{arXiv preprint arXiv:2206.04615}.

\bibitem[{Tayyar~Madabushi et~al.(2020)Tayyar~Madabushi, Romain, Divjak, and
  Milin}]{tayyar-madabushi-etal-2020-cxgbert}
Harish Tayyar~Madabushi, Laurence Romain, Dagmar Divjak, and Petar Milin. 2020.
\newblock \href {https://doi.org/10.18653/v1/2020.coling-main.355}
  {{C}x{GBERT}: {BERT} meets construction grammar}.
\newblock In \emph{Proceedings of the 28th International Conference on
  Computational Linguistics}, pages 4020--4032, Barcelona, Spain (Online).
  International Committee on Computational Linguistics.

\bibitem[{Tseng et~al.(2022)Tseng, Shih, Chen, Chou, Ku, and
  Hsieh}]{tseng-etal-2022-cxlm}
Yu-Hsiang Tseng, Cing-Fang Shih, Pin-Er Chen, Hsin-Yu Chou, Mao-Chang Ku, and
  Shu-Kai Hsieh. 2022.
\newblock \href {https://aclanthology.org/2022.lrec-1.683} {{C}x{LM}: A
  construction and context-aware language model}.
\newblock In \emph{Proceedings of the Thirteenth Language Resources and
  Evaluation Conference}, pages 6361--6369, Marseille, France. European
  Language Resources Association.

\bibitem[{Warstadt et~al.(2020)Warstadt, Zhang, Li, Liu, and
  Bowman}]{warstadt-etal-2020-learning}
Alex Warstadt, Yian Zhang, Xiaocheng Li, Haokun Liu, and Samuel~R. Bowman.
  2020.
\newblock \href {https://doi.org/10.18653/v1/2020.emnlp-main.16} {Learning
  which features matter: {R}o{BERT}a acquires a preference for linguistic
  generalizations (eventually)}.
\newblock In \emph{Proceedings of the 2020 Conference on Empirical Methods in
  Natural Language Processing (EMNLP)}, pages 217--235, Online. Association for
  Computational Linguistics.

\bibitem[{Wei et~al.(2021)Wei, Garrette, Linzen, and
  Pavlick}]{wei-etal-2021-frequency}
Jason Wei, Dan Garrette, Tal Linzen, and Ellie Pavlick. 2021.
\newblock \href {https://doi.org/10.18653/v1/2021.emnlp-main.72} {Frequency
  effects on syntactic rule learning in transformers}.
\newblock In \emph{Proceedings of the 2021 Conference on Empirical Methods in
  Natural Language Processing}, pages 932--948, Online and Punta Cana,
  Dominican Republic. Association for Computational Linguistics.

\bibitem[{Weissweiler et~al.(2022)Weissweiler, Hofmann, K{\"o}ksal, and
  Sch{\"u}tze}]{weissweiler-etal-2022-better}
Leonie Weissweiler, Valentin Hofmann, Abdullatif K{\"o}ksal, and Hinrich
  Sch{\"u}tze. 2022.
\newblock \href {https://aclanthology.org/2022.emnlp-main.746} {The better your
  syntax, the better your semantics? probing pretrained language models for the
  {E}nglish comparative correlative}.
\newblock In \emph{Proceedings of the 2022 Conference on Empirical Methods in
  Natural Language Processing}, pages 10859--10882, Abu Dhabi, United Arab
  Emirates. Association for Computational Linguistics.

\bibitem[{Wiedemann et~al.(2019)Wiedemann, Remus, Chawla, and
  Biemann}]{wiedemann2019}
Gregor Wiedemann, Steffen Remus, Avi Chawla, and Chris Biemann. 2019.
\newblock Does bert make any sense? interpretable word sense disambiguation
  with contextualized embeddings.
\newblock \emph{arXiv preprint arXiv:1909.10430}.

\bibitem[{Yaghoobzadeh et~al.(2019)Yaghoobzadeh, Kann, Hazen, Agirre, and
  Sch{\"u}tze}]{yaghoobzadeh-etal-2019-probing}
Yadollah Yaghoobzadeh, Katharina Kann, T.~J. Hazen, Eneko Agirre, and Hinrich
  Sch{\"u}tze. 2019.
\newblock \href {https://doi.org/10.18653/v1/P19-1574} {Probing for semantic
  classes: Diagnosing the meaning content of word embeddings}.
\newblock In \emph{Proceedings of the 57th Annual Meeting of the Association
  for Computational Linguistics}, pages 5740--5753, Florence, Italy.
  Association for Computational Linguistics.

\bibitem[{Zhan(2017)}]{zhan_2017}
Weidong Zhan. 2017.
\newblock On theoretical issues in building a knowledge database of chinese
  constructions.
\newblock \emph{Journal of Chinese Information Processing}, 31(1):230–238.

\bibitem[{Ziem et~al.(forthcoming)Ziem, Willich, and Michel}]{constructicon}
Alexander Ziem, Alexander Willich, and Sascha Michel. forthcoming.
\newblock \emph{Constructing constructicons}.
\newblock John Benjamins Publishing Company.

\end{thebibliography}
\bibliographystyle{acl_natbib}

\end{document}